\documentclass[runningheads]{llncs}

\usepackage{eccv}

\usepackage{eccvabbrv}

\usepackage{graphicx}
\usepackage{booktabs}
\usepackage{multirow}
\usepackage{enumitem}
\usepackage[table]{xcolor}

\newcommand{\best}[1]{\colorbox{red!20}{#1}}
\newcommand{\second}[1]{\colorbox{orange!20}{#1}}

\usepackage[accsupp]{axessibility}  

\usepackage{hyperref}

\usepackage{orcidlink}

\begin{document}

\title{LiteGS: A High-performance Framework to Train 3DGS in Subminutes via System and Algorithm Codesign} 

\titlerunning{LiteGS: A High-performance Framework to Train 3DGS}

\author{
Kaimin Liao \and
Hua Wang \and
Zhi Chen \and
Luchao Wang \and
Yaohua Tang 
}

\authorrunning{K.~Liao et al.}

\institute{
Moore Threads AI, Shenzhen, China\\
\email{tangyaohua28@gmail.com}
}

\maketitle

\begin{abstract}

  While 3D Gaussian Splatting (3DGS) has revolutionized novel view synthesis with its photorealistic and real-time rendering, its lengthy training process remains a critical bottleneck. Existing acceleration methods largely treat algorithm design and system optimization in isolation, failing to fully unleash hardware potential. To systematically address this, we propose \textit{LiteGS}, a high-performance framework driven by a fundamental system-algorithm codesign. At the GPU and data layers, LiteGS introduces a warp-based rasterization paradigm and a Cluster-Cull-Compact pipeline, fundamentally resolving data conflicts and spatial locality issues. Empowered by this robust backend, we introduce a variance-guided densification metric at the algorithm layer to accurately identify under-reconstructed regions. This powerful synergy establishes a new state-of-the-art: LiteGS consistently achieves the highest rendering quality and the fastest training speed across all parameter scales. Extensive evaluations demonstrate our milestone performance. Compared to the high-quality SOTA (3DGS-MCMC), LiteGS achieves at least a 10$\times$ speedup while matching or exceeding its fidelity. Furthermore, by unlocking a ``space-for-time'' aggressive densification strategy, LiteGS successfully compresses the training time to the sub-minute level on RTX3090 (further to $\sim$30 seconds on RTX 4090) while securing rendering quality comparable to the original 3DGS.
  
\end{abstract}

\section{Introduction}
\label{sec:inroduction}

\begin{figure}[t]
    \centering
    \begin{subfigure}[b]{0.39\linewidth}
        \includegraphics[width=\linewidth]{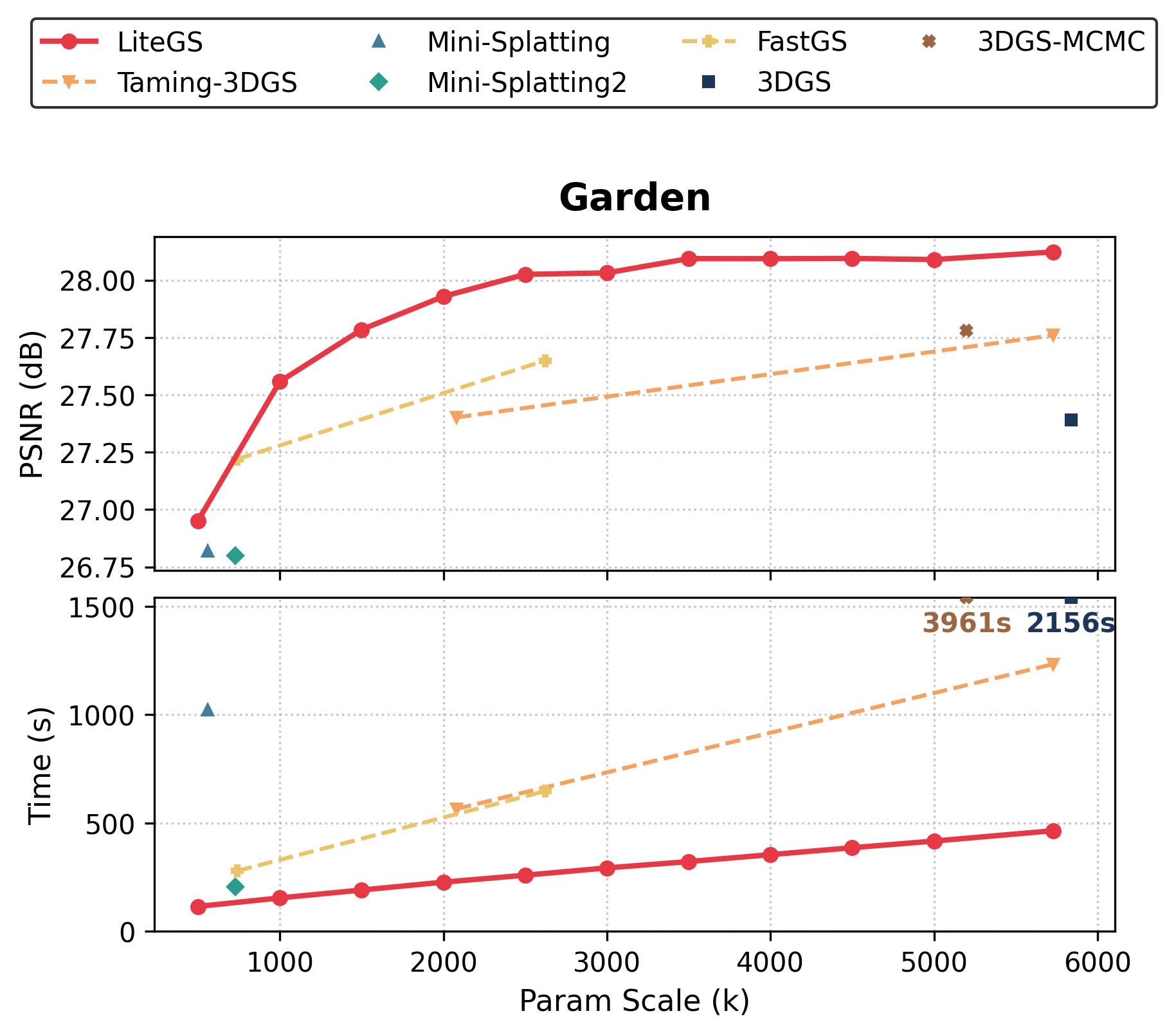}
    \end{subfigure}
    \hfill
    \begin{subfigure}[b]{0.6\linewidth}
        \includegraphics[width=\linewidth]{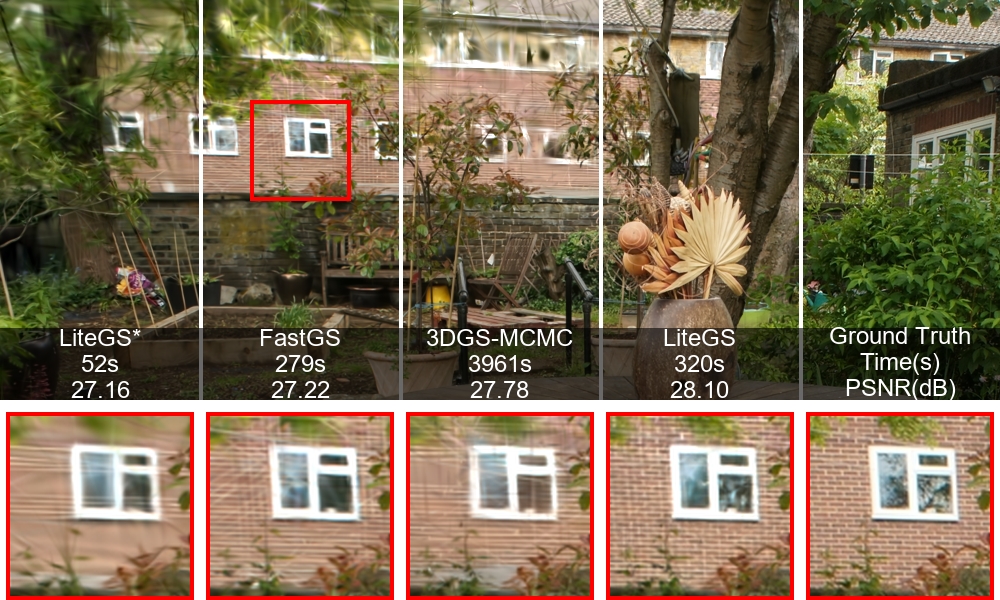}
    \end{subfigure}
    \caption{\textbf{Performance and visual comparisons of LiteGS against other works.} \textbf{Left:} Continuous evaluation on the \textit{Garden} scene. LiteGS consistently achieves the highest rendering quality (PSNR) across all parameter scales. More importantly, while the training time of baselines drastically inflates as parameters grow, LiteGS maintains an exceptionally low and flat time curve. \textbf{Right:} Qualitative comparisons. In the fast-reconstruction regime, our aggressive variant (LiteGS*) achieves sub-minute training (52s) while successfully recovering structural details. In the high-quality regime, LiteGS surpasses the quality SOTA (3DGS-MCMC) with a 12.3$\times$ speedup.}
    \label{image:curve}
\end{figure}

3DGS has achieved remarkable breakthroughs in novel view synthesis by representing scenes with millions of anisotropic Gaussians for photorealistic, real-time rendering~\cite{kerbl20233d}. This technique demonstrates immense potential across fields like autonomous driving and virtual reality~\cite{lin2024vast,liu2024citygaussian,vrsplat2025,radarsplat2025}. However, while rendering is exceptionally fast, training can take tens of minutes to hours, limiting its broader application.
Existing acceleration efforts~\cite{mallick2024taming,DISTWAR,TCGS} often target isolated components and fail to resolve deeper systemic bottlenecks. Specifically, gradient reduction suffers from severe data conflicts at the low-level GPU layer; disordered Gaussian storage causes poor spatial locality and cache misses at the data management layer; and conventional densification strategies lead to suboptimal geometric growth at the algorithm layer.

To systematically address these issues, we propose \textit{LiteGS}, an open-source, high-performance framework driven by a fundamental \textbf{system-algorithm codesign}. We move beyond isolated optimizations to comprehensively enhance the entire 3DGS training workflow:

\noindent \textbf{GPU Layer:} We propose a warp-based raster paradigm(one warp per tile) that simplifies intra-tile gradient aggregation into a single intra-warp reduction. This drastically reduces computational overhead while enabling efficient pixel-level statistics gathering.

\noindent \textbf{Data Management Layer:} We introduce a ``Cluster-Cull-Compact'' pipeline using Morton coding for ultra-fast dynamic spatial sorting. By culling and compacting memory at the cluster granularity, we radically improve spatial locality, reducing cache misses and warp divergence.

\noindent \textbf{Algorithm Layer:} Empowered by the statistical capabilities of our efficient rasterizer, we adopt the variance of per-pixel opacity gradients as a robust densification criterion. This consistently ensures optimal rendering quality at any parameter scale. Furthermore, fully supported by our high-throughput backend, it unlocks a ``space-for-time'' strategy---trading parameter efficiency for rapid geometric growth to achieve sub-minute training with competitive fidelity.

This co-optimization grants LiteGS unprecedented superiority, allowing it to consistently achieve the highest rendering quality and the fastest training speed across all parameter scales. Compared to the quality SOTA (3DGS-MCMC~\cite{kheradmand20243d}), LiteGS achieves at least a 10x speedup while matching or exceeding its fidelity. Furthermore, in the fast-reconstruction regime, by leveraging the ``space-for-time'' strategy, LiteGS successfully compresses the training time to the sub-minute level while securing rendering quality comparable to the original 3DGS~\cite{kerbl20233d}.

In summary, our main contributions are:
\begin{itemize}[leftmargin=*, noitemsep, topsep=0pt]
    \item We identify deep systemic bottlenecks in 3DGS training and introduce warp-based rasterization, Cluster-Cull-Compact, and variance-guided densification to systematically resolve them.
    \item We propose LiteGS, a high-performance framework built on system-algorithm codesign, establishing a new state-of-the-art by dominating both training efficiency and rendering quality across all parameter scales.
    \item We provide extensive evaluations demonstrating LiteGS's milestone performance: achieving at least a 10x speedup over the high-quality SOTA, and successfully unlocking a sub-minute training regime for fast reconstruction(reaching 30-40 seconds on modern high-end GPUs).
\end{itemize}

\section{Related Work}

\subsection{Efficient and Accelerated 3DGS} 

Although 3DGS is already faster to optimize than classic NeRF, the training still takes tens of minutes or even hours, and its cost grows quickly with the number of Gaussian primitives. Recent works improve efficiency from different perspectives. 

One line of work reduces the number of primitives or primitive--pixel interactions. Several works reduce the per-pixel Gaussian overhead by culling or skipping negligible contributions~\cite{Liu2025,Hanson2025b,Feng2025}. LightGaussian~\cite{lightgaussian} prunes redundant Gaussians with compactness-oriented heuristics, while PUP 3D-GS~\cite{Hanson2025a} uses a second-order sensitivity score to remove a large fraction of Gaussians from pretrained models while preserving visual fidelity. Speedy-Splat~\cite{hanson2025speedy} reduces unnecessary rasterization work by computing tighter and more accurate Gaussian--tile intersections with SnugBox and AccuTile.

Another line of work accelerates optimization by scheduling the training complexity. DashGaussian~\cite{chen2025dashgaussian} progressively increases the rendering resolution and synchronizes primitive growth with the resolution schedule, thereby avoiding redundant high-resolution computation in early training.  Mini-Splatting2~\cite{fang2024mini2} uses visibility information to reduce redundant computation.

Orthogonal to these algorithm-level acceleration strategies, several works improve the GPU utilization of the 3DGS rasterization pipeline. DISTWAR~\cite{DISTWAR} reduces atomic-update overhead in gradient accumulation with warp-level reduction. Taming 3DGS~\cite{mallick2024taming} proposes an alternative per-splat backpropagation strategy and several numerically equivalent or quality-preserving approximations to reduce training cost under a fixed Gaussian budget. TC-GS~\cite{TCGS} maps parts of the splatting computation to Tensor Cores by reformulating alpha blending as matrix operations. In contrast to these isolated optimizations, LiteGS jointly addresses rasterization, memory locality, and densification, so that the system backend and the model-growth strategy reinforce each other.

\subsection{Densification Strategies} 

A key to 3DGS's success is its ability to refine the point cloud during training via Adaptive Density Control (ADC), i.e., adding Gaussians in under-reconstructed regions and removing those with negligible opacity. However, the original ADC heuristic relies heavily on screen-space positional gradients and opacity thresholds, which may create redundant primitives or miss regions where the gradient magnitude is ambiguous. Recent works therefore revisit densification to make model growth more principled and efficient. Bul{`o} \etal~\cite{Revising3DGS} propose an error-driven formulation that uses an auxiliary per-pixel error map to guide where new Gaussians should be created. 3DGS-MCMC~\cite{kheradmand20243d} reformulates densification and pruning as deterministic state transitions in an MCMC framework, replacing hand-crafted clone/split heuristics with probability-preserving Gaussian relocation. ConeGS~\cite{baranowski2026conegs} creates new primitives from high-error pixels by placing cone-sized Gaussians along estimated pixel viewing cones, reducing the reliance on random splits from existing primitives.

Other methods control when and how many Gaussians should be added. Taming 3DGS~\cite{mallick2024taming} uses score-based constructive densification to grow the model toward a predetermined budget, adding Gaussians that are estimated to improve reconstruction quality the most. Mini-Splatting~\cite{fang2024mini} improves compactness by combining densification with simplification, while Mini-Splatting2~\cite{fang2024mini2} shows that aggressive early densification can accelerate geometry formation when paired with later simplification. FastGS~\cite{ren2025fastgs} uses multi-view consistency to guide both densification and pruning, avoiding uncontrolled Gaussian growth. Unlike these methods, LiteGS uses the variance of per-pixel opacity gradients as a full-view densification signal. This criterion distinguishes consistent optimization demand from conflicting pixel-level evidence, and is made efficient by our warp-based rasterizer.

\section{Background: 3D Gaussian Splatting}
\label{sec:background}

3DGS training starts with a point cloud initialized from Structure-from-Motion (SFM), followed by the optimization of primitive parameters via gradient descent. To enhance reconstruction fidelity, the framework strategically performs densification at regular intervals, augmenting the number of primitives. The Gaussian scene is represented as $\mathbb{G} = \{ g_i | i = 1, \ldots, N \}$, a collection of $N$ primitives. Each primitive $g_i = \{ \mathbf{p}^{\text{world}}_i, \mathbf{\Sigma}^{\text{world}}_i, \mathbf{c}_i, o_i \}$ is defined by its 3D position in world coordinates $\mathbf{p}_{\text{world}}^i$, a 3D covariance matrix $\mathbf{\Sigma}_{\text{world}}^i$, RGB color coefficients $\mathbf{c}_i$, and an opacity value $o_i$. The camera views are donated as $\Pi=\{\pi_i|i=1,\ldots, M\}$ and $M$ is the number of views in the training set. The Gaussian primitive projected into screen space is represented as $\{x_{i,\pi_j}^{screen},y_{i,\pi_j}^{screen},\Sigma_{i,\pi_j}^{screen}\,\mathbf{c}_{i}, o_i \}$. The rendering process in 3DGS is mathematically defined as, 

\begin{equation}
\label{equ:render}
\mathbf{I}_{\pi_{j}}(x,y) = \sum_{i=0}^{N'} T_{s(i),\pi_{j}}(x,y) \cdot \alpha_{s(i),\pi_{j}}(x,y) \cdot \mathbf{c}_{s(i)}
\end{equation}
where \(s(i)\) denotes the index mapping function for depth-ordered primitives in camera space, corresponding to the \(i\)-th sorted primitive in the Gaussian scene representation \(\mathbb{G}\), $T$ is the transmittance $T_{s(i),\pi_{j}}(x,y)=T_{s(i-1),\pi_j}(x,y) \cdot (1-\alpha_{s(i-1),\pi_j}(x,y))$\normalsize, \(N'\) represents the number of primitives visible within the view frustum,   $\alpha$ is the opacity term $\alpha_{i,\pi_j}(x,y)=o_i G(x_{i,\pi_j}^{screen}-x,y_{i,\pi_j}^{screen}-y,\Sigma_{i,\pi_j}^{screen})$ \normalsize and $G$ is defined as,  

\begin{equation}
\label{equ:gaussian}
G(\Delta x, \Delta y, \Sigma) = e^{-0.5 (\Delta x, \Delta y) \Sigma^{-1} (\Delta x, \Delta y)^T}
\end{equation}

We use $\mathbb{P}=\{P_j|j=1,...,N_{tile}\}$ to represent the $j$-th rendering tiles, and the gradient of $c^{r}_{s(i)}$ is computed by \cref{equ:gradient}.
\begin{equation}
\label{equ:gradient}
\frac{\partial Loss}{\partial c^{r}_{s(i)}}=\sum_{j=0}^{N_{tile}}\sum_{(x,y) \in P_{j}}\frac{\partial Loss}{\partial I_{r}(x,y)} \cdot T_{s(i),\pi}(x,y) \cdot \alpha_{s(i),\pi}(x,y)
\end{equation}

\begin{figure}[t]
    \centering
    \includegraphics[width=\textwidth]{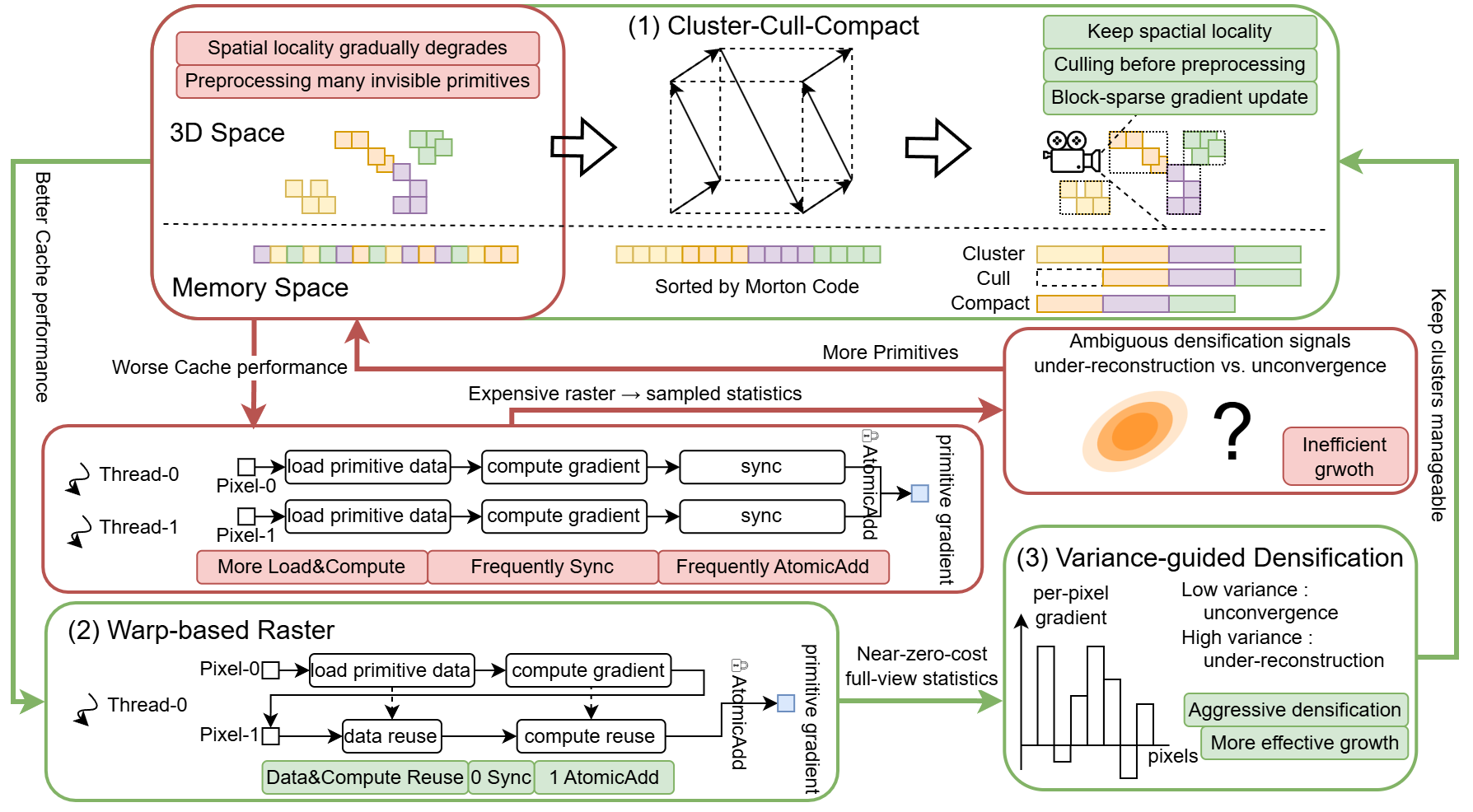}
    \caption{LiteGS converts the negative cross-layer loop of conventional 3DGS training into a positive system–algorithm co-design. Red denotes bottlenecks in conventional 3DGS; green denotes the corresponding LiteGS designs. In conventional pipelines, ambiguous densification cues may introduce ineffective primitives. Besides, the appending primitives degrades spatial locality and increases preprocessing and rasterization overhead. The inefficient rasterizer further makes dense pixel-level statistics expensive, forcing densification to rely on primitive-level or sampled cues and thus perpetuating ambiguous growth decisions. LiteGS breaks this loop with three coupled modules: (1) Cluster-Cull-Compact restores locality and culls invisible clusters early. (2) Warp-based Rasterization reduces synchronization and atomic accumulation while collecting pixel-level statistics at near-zero overhead. (3) Variance-guided Densification uses opacity-gradient variance to make more reliable split decisions. Together, these modules reinforce each other and make a faster and better training. }
    \label{fig:codesign_overview}
\end{figure}

\section{Methodology}

While the theoretical formulation of 3DGS is elegant (\cref{sec:background}), its practical training pipeline suffers from severe systemic degradation at scale. These bottlenecks are not isolated; they stem from tightly coupled inefficiencies across low-level computation (atomic conflicts), mid-level memory management (disordered layout), and top-level algorithmic design (ambiguous densification metrics).

To break this systemic deadlock, we propose \textbf{LiteGS}, a high-performance framework built upon a \textit{System-Algorithm Codesign} philosophy. As illustrated in Fig.~\ref{fig:codesign_overview}, LiteGS synergistically tackles these challenges through three integrated modules:

\begin{itemize}[leftmargin=*, noitemsep, topsep=0pt]
    \item \textbf{Warp-based Rasterization} (\cref{subsec:raster}): Moving beyond incremental patches, we propose a fundamentally new parallel execution model that comprehensively eliminates the underlying computation stalls. Moreover, it unlocks zero-overhead, pixel-level statistical tracking to empower our densification strategy.
    
    \item \textbf{Cluster-Cull-Compact} (\cref{subsec:cull}): A pipeline that introduces an early cluster-level culling scheme. Furthermore, it dynamically enforces spatial locality, yielding global benefits by ensuring coalesced memory access for the entire pipeline, including our newly designed warp-based rasterizer.
    
    \item \textbf{Variance-guided Densification} (\cref{subsec:densify}): Exploiting the ``free'' statistics provided by our rasterizer, we establish a more robust densification metric. This criterion successfully overcomes the fundamental blind spots of prior heuristic methods.
\end{itemize}

\subsection{Warp-based Raster}
\label{subsec:raster}

The backward rasterization of 3DGS has an inherent many-to-one reduction structure. The gradient of a Gaussian primitive is accumulated from all pixels covered by that primitive. A conventional pixel-parallel schedule is therefore mismatched with the backward pass: different threads process different pixels efficiently, but their partial gradients eventually compete to update the same primitive-level buffers. This mismatch leads to three structural inefficiencies. First, many pixels may atomically update the same Gaussian gradient. Second, neighboring pixels repeatedly load the same Gaussian attributes. Third, the common block-per-tile mapping requires explicit synchronization when threads cooperatively process primitives within a tile.

Prior works have recognized parts of this bottleneck. For example, DISTWAR~\cite{DISTWAR} reduces atomic-update overhead with warp-level reduction, and Taming 3DGS~\cite{mallick2024taming} changes the backpropagation granularity to avoid some tile-level conflicts. These methods confirm that rasterization is a major training hotspot. However, our profiling shows that the inefficiency of the original backward rasterizer is not caused by a single symptom. Instead, low resident parallelism and multiple stall families jointly limit GPU utilization.

\begin{figure}[b]
    \centering
    \begin{subfigure}[b]{0.49\linewidth}
        \includegraphics[width=\linewidth]{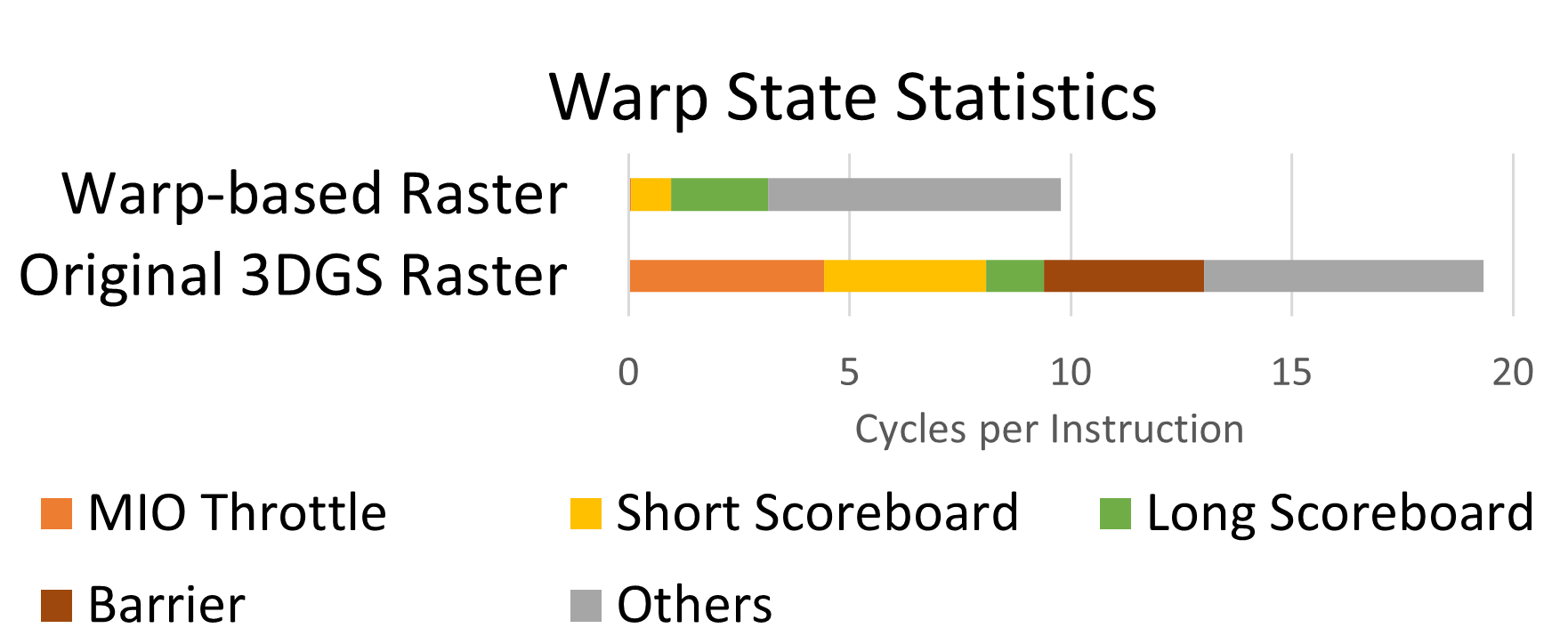}
    \end{subfigure}
    \hfill
    \begin{subfigure}[b]{0.49\linewidth}
        \includegraphics[width=\linewidth]{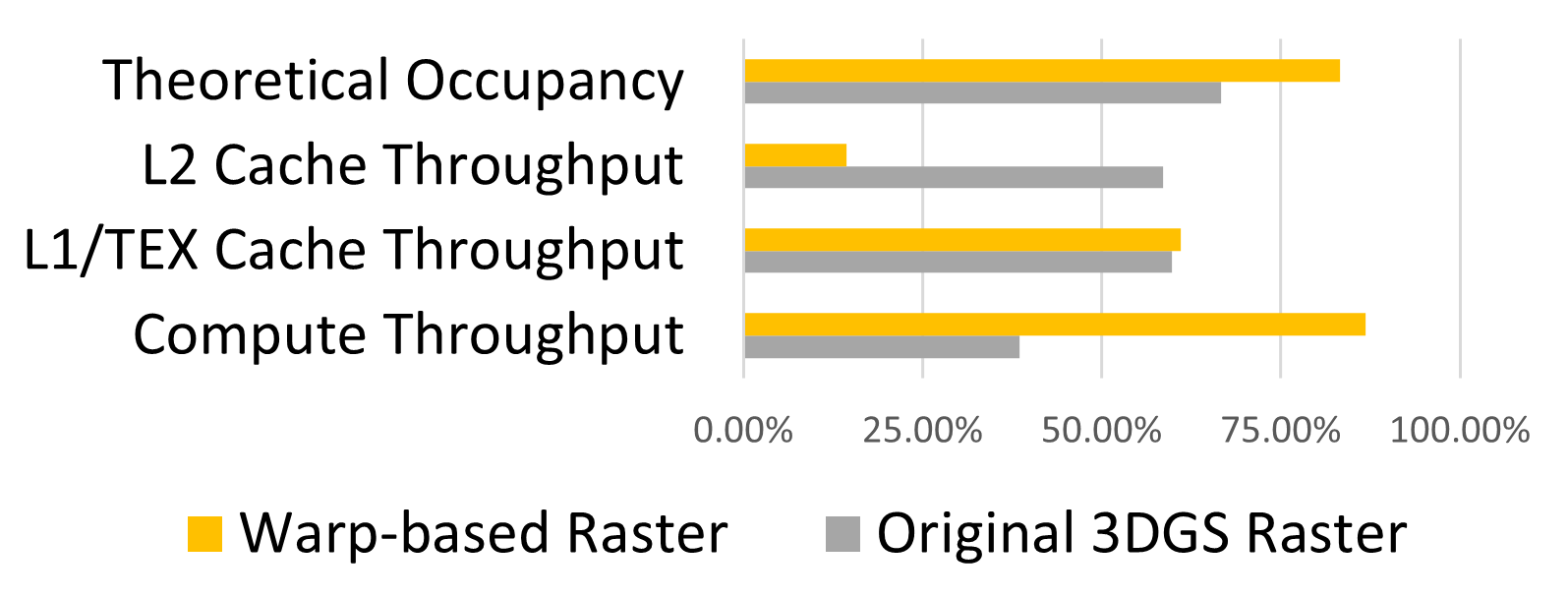}
    \end{subfigure}
    \caption{\textbf{Profiling-guided diagnosis of backward rasterization.} The original 3DGS backward rasterizer is not limited by a single bottleneck. It suffers from low compute throughput together with multiple stall families, including atomic-pipeline congestion (MIO throttle), memory-dependency stalls (short/long scoreboard), and block-level synchronization stalls (barrier). Our warp-based rasterizer changes the execution model from block-level pixel-parallel processing to warp-local gradient aggregation. This substantially increases compute throughput.}
    \label{image:perf}
\end{figure}

Fig.~\ref{image:perf} summarizes this diagnosis. Occupancy measures how many warps can reside on an SM(Streaming multiprocessor) and hide latency, while stall cycles measure why warps cannot issue its next instruction. The higher warp cycles per instruction, the more warp parallelism is required to hide this latency. The original 3DGS rasterizer has only moderate theoretical occupancy and serious stalls, resulting in low compute throughput and low memory throughput. Its stall breakdown shows substantial atomic-pipeline congestion (MIO throttle), memory-dependency stalls (short/long scoreboard), and block-level synchronization stalls (barrier). This explains why optimizing only atomic accumulation is insufficient: even if one stall source is reduced, the kernel can remain limited by memory dependency, synchronization, or insufficient occupancy. 

Therefore, efficient 3DGS training requires more than patching individual stalls; it demands a hardware-aware rasterization operator co-designed with the gradient reduction, I/O, and synchronization patterns of 3DGS backpropagation. Our key idea is to make a warp, rather than a thread block or an individual pixel thread, the basic execution unit of a tile. This leads to \textbf{Warp-based Rasterization}, where each warp owns one tile, each lane processes multiple neighboring pixels, and primitive gradients are locally accumulated before being written back to global memory. This execution model simultaneously addresses the three bottlenecks identified above:

\begin{itemize}[leftmargin=*, noitemsep, topsep=0pt]
    \item \textbf{Local accumulation before global update:} 
    Traditional pixel-centric mappings require multiple threads to concurrently issue global \textit{atomicAdd} operations for the same Gaussian primitive, leading to severe memory pipeline congestion. We shift the paradigm by having a single thread handle multiple pixels. Gradients are first accumulated within local registers (\textit{intra-thread reduction}). This ensures that pixel-level conflicts are resolved at the register level, minimizing the required global atomic operations.
    
    \item \textbf{Gaussian reuse across neighboring pixels:} 
    By assigning multiple pixels to a single thread, once a Gaussian's attributes (e.g., covariance, color) are loaded into the registers, they can be reused across all assigned pixels. This multipixel-per-thread design significantly reduces redundant memory fetch instructions. \textit{Furthermore, even the initial loading of Gaussian parameters by each thread significantly benefits from the spatial contiguity maintained by the Cluster-Cull-Compact pipeline (\cref{subsec:cull}).}
    
    \item \textbf{Keep each tile inside one warp:} 
    The conventional parallel model maps a thread block to a rendering tile. Processing primitives cooperatively within a tile inevitably requires frequent intra-block synchronization to ensure data consistency among threads. We break this dependency by establishing a strict ``one warp per tile'' principle. By confining the entire workload of a tile to a single warp (32 threads), our execution model naturally operates in lockstep at the warp level and exploits hardware-native shuffle instructions (e.g., \texttt{\_\_shfl\_down\_sync}) for data exchange. This architectural shift fundamentally removes the need for block-level synchronization.
\end{itemize}

A critical design choice in our new parallel model is determining the optimal number of pixels processed by a single thread. Processing more pixels per thread requires allocating more local registers to save intermediate blending variables(offloading these to shared memory incurs unacceptable latency). However, excessive register consumption per thread reduces theoretical occupancy, thereby reducing overall hardware throughput. 

Ideally, to maximize performance across various architectures, one should establish a hardware-aware configuration: first, determine the maximum pixels-per-thread ratio based on the per-pixel register usage and the SM's register file size to maintain high theoretical occupancy; then, derive the optimal tile size following the strict one-warp-per-tile principle.

Unfortunately, assigning multiple pixels per thread exacerbates the baseline's already severe register pressure (capped at a mere 66\% theoretical occupancy for a single pixel). We introduce two key techniques to reduce the per-pixel register usage: \textbf{Forward Differencing} and \textbf{Mixed-Precision}.

\subsubsection{Forward Differencing} \label{sec:forward_diff}

This classical numerical optimization widely adopted in early hardware and software rasterizers~\cite{lien1987adaptive, heckbert1989fundamentals,zwicker2001ewa}. We adapt this mathematical paradigm to our warp-based model.

This idea stems from the observation that when processing multiple consecutive pixels for the same Gaussian primitive along a scanline (e.g., the y-axis), the exponent of the Gaussian function (\cref{equ:gaussian}) can be decomposed into a quadratic polynomial of the pixel step offset $j$: 
$E(j) = \text{Basic} + \text{Linear} \cdot j + \text{Quad} \cdot j^2$. Rather than evaluating this polynomial directly per pixel, we apply \textit{forward differencing} to update the exponent iteratively:
\begin{equation}
\begin{aligned}
    E_{j+1} &= E_{j} + \Delta E_{j} \\
    \Delta E_{j+1} &= \Delta E_{j} + \Delta^2 E
\end{aligned}
\end{equation}
where $\Delta^2 E = 2 \cdot \text{Quad}$.

This algorithmic reformulation yields two profound hardware-level benefits:
\begin{enumerate}
    \item \textbf{Register Compression:} It effectively compresses the geometric context needed for each pixel calculation from 7 variables( $x$, $x^{\text{screen}}_i$ , $y$, $y^{\text{screen}}_i$, $(\Sigma^{-1})_{11}$,$(\Sigma^{-1})_{12}$,$(\Sigma^{-1})_{22}$) down to merely 3 running variables ($E$, $\Delta E$, $\Delta^2 E
    $).
    \item \textbf{Less Computation:} Beyond the initial setup, evaluating the Gaussian exponent for all subsequent pixels requires only two simple incremental updates, significantly reducing arithmetic instructions in the inner loop.
\end{enumerate}

Crucially, this structural benefit extends to the backward pass. 

\subsubsection{Mixed-Precision Computing}

The second technique leverages mixed-precision to further reduce register pressure. For alpha blending, we strategically select half-precision (FP16) instead of single-precision (FP32) for temporary variables during the blending process (e.g. Gaussian intensity $G(\Delta x, \Delta y, \Sigma^{\text{screen}})$, transmittance $T$, and color $c$).

\subsection{Cluster-Cull-Compact} 
\label{subsec:cull}

As training progresses, 3DGS frequently appends newly generated Gaussian primitives to the end of memory arrays. This unordered layout destroys spatial locality, causing severe cache misses during memory accesses and massive warp divergence during visibility culling.

To address this disordered memory issue without introducing heavy overhead, we adapt the established ``Cluster-Cull-Compact'' pipeline~\cite{karis2021nanite} to the dynamic training loop of 3DGS, as illustrated in the bottom-left of \cref{fig:codesign_overview}. Rather than building complex bounding volume hierarchies (BVH) which are too slow for per-iteration updates, our pipeline enforces spatial locality through lightweight Morton coding. 

The pipeline consists of three efficient stages:
\begin{enumerate}
    \item \textbf{Morton-based Sorting:} We map the 3D coordinates of Gaussians to a 1D sequence using 64-bit Morton codes. By periodically sorting the primitive array based on these codes (e.g., after each densification step), we quickly restore spatial locality. The execution time of this sorting is minimal, making the overhead negligible.
    \item \textbf{Cluster-level Culling:} Based on the sorted array, we group contiguous primitives into fixed-size clusters (e.g., 128 primitives per cluster) and compute an axis-aligned bounding box (AABB) for each. View frustum culling is then performed directly against these AABBs. This early cluster-level culling allows us to discard massive numbers of invisible primitives with minimal computation, significantly accelerating the preprocessing phase.
    \item \textbf{Compaction and Synergy:} Finally, the data of all visible clusters are compacted into a continuous memory buffer. This continuous layout is critical: it guarantees coalesced memory access for our warp-based rasterizer (\cref{subsec:raster}), effectively hiding the memory latency. 
\end{enumerate}

\paragraph{Synergy with Sparse Optimizer Updates.}
Furthermore, this spatially-ordered cluster structure naturally benefits sparse gradient update strategies. Instead of applying parameter updates (e.g., sparse Adam) at the scattered per-primitive level, we apply sparsity at the cluster granularity. This ensures that the optimizer always reads and writes contiguous memory blocks, thereby maximizing memory bandwidth utilization and avoiding the performance penalties associated with random memory accesses.

\subsection{Variance-guided Densification}
\label{subsec:densify}

The primary goal of densification is to adaptively add geometry in under-fitted regions. Previous works typically rely on the magnitude of positional gradients or pixel-space errors to identify these areas. However, these magnitude-based metrics suffer from a fundamental ambiguity: a large gradient or error might indicate a genuine under-fitted region requiring new primitives, or it could merely represent a well-sampled but temporarily unconverged Gaussian. This ambiguity is particularly pronounced during early training stages or immediately following periodic opacity resets.

To definitively resolve this ambiguity, we introduce a robust metric based on the variance of the opacity gradient, as illustrated in the bottom-right of \cref{fig:codesign_overview}. Instead of evaluating the aggregated magnitude, we analyze the spatial distribution of gradients across all pixels observing the same primitive:
\begin{itemize}[leftmargin=*, noitemsep, topsep=0pt]
    \item \textbf{Low Variance:} If opacity gradients are large but directionally consistent, the optimizer has reached a consensus. The primitive is simply unconverged; no new geometry is needed.
    \item \textbf{High Variance:} If gradients vary significantly among pixels (e.g., conflicting visibility demands at occlusion boundaries), this spatial conflict signals a true geometric deficiency, making the primitive a prime candidate for splitting.
\end{itemize}

To prevent periodic opacity resets from inducing sudden transient errors that confuse our metric, we replace the traditional hard reset with a gentler ``opacity decay'' (halving the opacity instead of zeroing it).

Furthermore, aggregating pixel-level information across the entire image is traditionally time-consuming, forcing pixel-error-based methods to rely on sparse view sampling to maintain training speed. In contrast, our warp-based rasterizer (\cref{subsec:raster}) naturally unlocks the ability to compute these pixel-level statistics (e.g., the sum of gradients and sum of squared gradients) with \textbf{zero overhead}. We can obtain full-view variance metrics without any computational penalty.

\subsubsection{Space-for-Time: Aggressive Densification}

Inspired by insights that primitives frequently stall in optimization saddle points~\cite{wang2025steepest}, we adopt a ``space-for-time'' aggressive densification strategy. Unlike traditional pipelines that restrict splitting to avoid parameter explosion, we forcefully split all parameters residing in saddle points, regardless of whether gradient descent might eventually pull them out. Within our highly optimized system, the training throughput is largely insensitive to the parameter scale, which safely unlocks this trade-off space. By aggressively splitting primitives, we effectively accelerate the convergence around saddle points, allowing us to achieve high-quality rendering in significantly fewer training iterations and boosting the overall training speed.

Recent works like Mini-Splatting2~\cite{fang2024mini2} also explore aggressive densification, but their utilization is fundamentally constrained. Lacking precise variance guidance and bottlenecked by unoptimized pipelines, they treat aggressive splitting merely as a temporary greedy strategy. Consequently, they waste nearly half their training iterations on expensive pruning and simplification to manage the inflated parameter count.

\section{Experiments}

We conducted extensive experiments to validate the effectiveness of our proposed framework. 

\noindent \textbf{Datasets}: We evaluated LiteGS on three datasets: Mip-NeRF 360~\cite{barron2022mip}, Tanks \& Temples~\cite{knapitsch2017tanks}, and Deep Blending~\cite{hedman2018deep}. 

\noindent\textbf{Baselines}: We compare LiteGS against a comprehensive set of baselines, including foundational methods like the original 3DGS~\cite{kerbl20233d}, high-quality reconstruction methods like 3DGS-MCMC~\cite{kheradmand20243d}, and recent acceleration-focused works such as Taming-3DGS~\cite{mallick2024taming}, DashGaussian~\cite{chen2025dashgaussian}, Mini-Splatting~\cite{fang2024mini}, Mini-Splatting2~\cite{fang2024mini2} and FastGS~\cite{ren2025fastgs}.

\noindent \textbf{Implementation Details}: All experiments are performed on a single NVIDIA RTX 3090 GPU. LiteGS adopts the same learning rate schedules as Taming-3DGS~\cite{mallick2024taming}. We evaluate our framework under two distinct training schedules:
\begin{itemize}[leftmargin=*, noitemsep, topsep=0pt]
    \item \textbf{LiteGS}: The default configuration, trained for a total of 30K iterations. Densification starts after the third epoch, with splitting operations performed every 5 epochs during the first 80\% of the training. Opacity decay is applied every 10 epochs.
    \item \textbf{LiteGS*}: Employs our aggressive densification strategy for extreme acceleration. Splitting operations are performed every 2 epochs, and the total training is significantly shortened to 10K iterations.
\end{itemize}

\noindent \textbf{Parameter Scale}: Similar to Taming-3DGS~\cite{mallick2024taming}, the parameter scale of LiteGS is highly configurable. We select two specific configurations to compare against fast-reconstruction and high-quality baselines, respectively. Since perfectly aligning parameter counts with all baselines is impractical due to their varying capacities, we additionally provide a comprehensive evaluation across the full spectrum of parameter scales. This ensures our rendering advantages are not merely the result of parameter inflation.

\subsection{Comparison with State-of-the-Art Methods}
\cref{tab:main_results} presents the quantitative comparison against state-of-the-art methods across all evaluated datasets. We note that the aggregated metrics for 3DGS-MCMC differ slightly from their original publication. This is because their officially reported results omitted two scenes from the Mip-NeRF 360 dataset (which were temporarily unavailable at the time of their evaluation). 

\begin{table}
\centering
\setlength{\tabcolsep}{1.5pt}
\renewcommand{\arraystretch}{1.1}
\setlength{\fboxsep}{0pt}
\scriptsize

\caption{Quantitative comparison on Mip-NeRF 360, Tanks\&Temples, and Deep Blending datasets. We report SSIM, PSNR, LPIPS, training time (T, in seconds), and the number of Gaussians (N, in millions). To accommodate page limits, some baselines are abbreviated (MS: Mini-Splatting\cite{fang2024mini} \cite{fang2024mini2}, MCMC: 3DGS-MCMC\cite{kheradmand20243d}, Taming: Taming-3DGS \cite{mallick2024taming}, Dash: DashGaussian\cite{chen2025dashgaussian}). Methods marked with * (e.g., MSv2*) employ an Aggressive Densification (AD) strategy with a reduced iteration budget.}
\label{tab:main_results}

\begin{tabular}{l|c|c|c}
\toprule
\multirow{2}{*}{\textbf{Method}} & \textbf{Mip-NeRF 360} & \textbf{Tanks\&Temples} & \textbf{Deep Blending} \\
\cmidrule(lr){2-2} \cmidrule(lr){3-3} \cmidrule(lr){4-4}
&SSIM/PSNR/LPIPS/T/N
&SSIM/PSNR/LPIPS/T/N
&SSIM/PSNR/LPIPS/T/N \\
\midrule

Taming
&0.801/27.32/0.257/ 380/0.68
&0.834/23.73/0.210/ 238/0.31
&0.904/29.90/0.255/ 294/\second{0.25}\\

MS
&\best{0.822}/27.32/\second{0.217}/1221/0.49
&\best{0.846}/23.43/\best{0.180}/ 755/\second{0.30}
&\second{0.910}/29.98/\best{0.240}/1036/0.56\\

MSv2*
&\second{0.821}/27.33/\best{0.215}/ 214/0.62
&\second{0.841}/23.14/0.186/ 142/0.35
&\best{0.912}/\second{30.08}/\best{0.240}/ 165/0.65\\

FastGS
& 0.797/27.56/0.261/ 215/\best{0.40}
& 0.839/\second{24.15}/0.210/ 141/\best{0.24}
& 0.901/30.03/0.270/ 143/\best{0.22} \\

LiteGS
& 0.804/\best{27.62}/0.250/\second{ 127}/\second{0.42}
& 0.840/\best{24.30}/0.200/\second{ 109}/\best{0.24}
& 0.904/30.01/0.268/\second{ 99 }/\best{0.22}\\

LiteGS*
& 0.813/\second{27.60}/0.229/\best{ 64 }/1.00 
& \second{0.841}/23.71/\second{0.184}/\best{ 67 }/1.00 
& 0.908/\best{30.14}/\second{0.252}/\best{ 47 }/1.00 \\
\midrule

3DGS
& 0.815/27.47/0.216/1626/3.35 
& 0.848/23.66/0.176/ 906/1.84 
& 0.904/29.54/0.244/1491/2.82 \\

Dash
& 0.816/27.66/0.220/ 412/2.23
& 0.851/24.02/0.180/ 318/1.20 
& \second{0.908}/\second{30.14}/0.247/ 285/\second{1.91} \\

Taming
& 0.822/27.84/\second{0.207}/ 906/3.30
& 0.855/24.17/\second{0.167}/ 547/1.83 
& \second{0.908}/29.88/\best{0.234}/ 813/2.80 \\

MCMC
& \best{0.834}/\second{28.08}/\best{0.186}/3142/3.21
& \second{0.860}/24.29/0.190/1506/1.85
& 0.890/29.67/0.320/2319/2.95\\

FastGS
& 0.820/27.93/0.216/\second{ 375}/\second{1.15}
& 0.855/\second{24.39}/0.175/\second{ 214}/\best{0.54}
& 0.907/30.12/0.243/\second{ 220}/\best{0.65}\\

LiteGS
& \second{0.827}/\best{28.14}/\second{0.207}/\best{ 194}/\best{1.11}
& \best{0.861}/\best{24.61}/\best{0.161}/\best{ 150}/\second{0.60}
& \best{0.911}/\best{30.36}/\second{0.241}/\best{ 128}/\best{0.65}\\

\bottomrule
\end{tabular}

\end{table}

As demonstrated in the top half of \cref{tab:main_results}, LiteGS completely dominates the lightweight and fast-reconstruction regime. Under standard configurations, LiteGS outperforms recent fast variants like FastGS in rendering quality (e.g., 27.62 vs. 27.56 PSNR on Mip-NeRF 360) while being nearly $1.7\times$ faster (127s vs. 215s). Furthermore, our aggressive variant, LiteGS*, successfully unlocks the sub-minute training milestone. It reaches convergence in just 64s, 67s, and 47s across the three respective datasets. It drastically outperforms MSv2* and FastGS in absolute speed while maintaining highly competitive rendering fidelity (e.g., matching or exceeding the PSNR of FastGS on Deep Blending and Mip-NeRF 360). While our primary evaluations use an RTX 3090 to ensure fair comparison with prior arts. When deployed on a high-end GPU(RTX 4090) the training time of LiteGS* is further compressed to just 40s on Mip-NeRF 360, 45s on Tanks\&Temples, and 29s on Deep Blending, establishing a new absolute speed record for high-fidelity 3DGS training.

In the high-quality regime (bottom half of \cref{tab:main_results}), LiteGS demonstrates unprecedented efficiency and parameter utilization. Most notably, when evaluated against the quality SOTA, 3DGS-MCMC, LiteGS not only consistently surpasses its rendering fidelity (e.g., $+0.32$ dB on Tanks\&Temples and $+0.69$ dB on Deep Blending) but also achieves a $10\times$ to $18\times$ speedup (e.g., 128s vs. 2319s on Deep Blending). Crucially, LiteGS accomplishes this high-fidelity reconstruction utilizing only about one-third of the parameters required by 3DGS-MCMC, rigorously validating the architectural superiority of our system-algorithm codesign.

\begin{figure}[t]
    \centering
    \includegraphics[width=\textwidth]{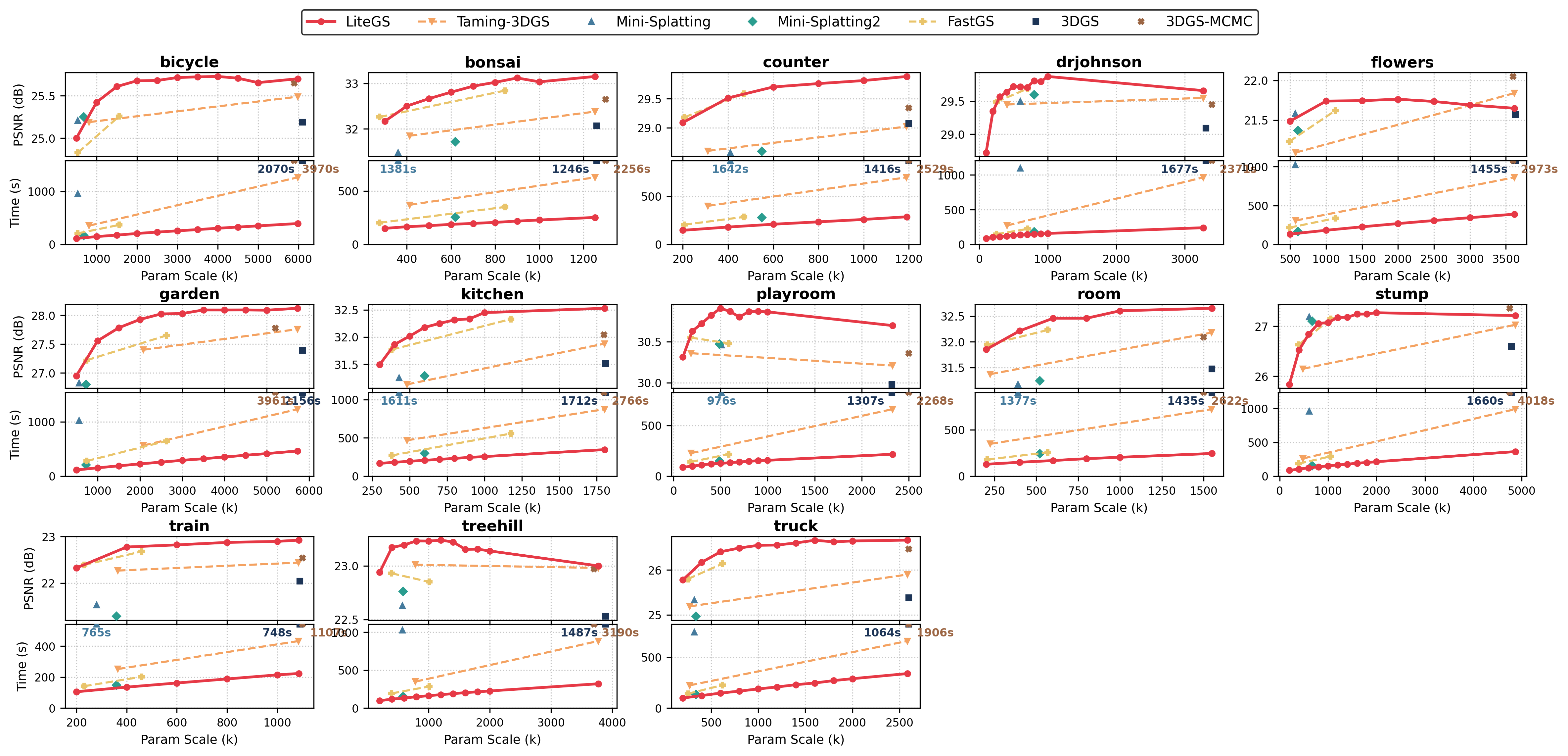}
    \caption{\textbf{Continuous evaluation across varying parameter scales.} We compare the rendering quality (PSNR) and training time against the number of primitives (Param Scale). Across diverse scenes, LiteGS consistently establishes the optimal Pareto front, achieving the highest PSNR and the fastest convergence at any given capacity budget. }
    \label{fig:param_scale}
\end{figure}

While \cref{tab:main_results} demonstrates our superiority at specific parameter budgets, a thorough evaluation requires analyzing performance across varying model capacities. To ensure a strictly fair comparison, we present a continuous evaluation of LiteGS against all baselines across a wide range of parameter scales in \cref{fig:param_scale}. As depicted in the figure, LiteGS consistently achieves the highest rendering quality and the fastest training speed across almost all parameter settings. It is particularly worth noting that some existing approaches achieve their reported trade-offs by employing entirely different training hyperparameters for "fast" versus "high-quality" targets, or even by relying on per-scene hyperparameter tuning. In stark contrast, LiteGS maintains a strictly unified configuration across all scenes and all quality targets.

\subsection{Ablation Studies}
This section validates the effectiveness of each key component of LiteGS through ablation studies.

\noindent\textbf{Effectiveness of Aggressive Densification} To validate that our aggressive densification (AD) benefits from system-level optimization rather than simple hyperparameter tuning, we compare LiteGS with several baselines under a 10K-iteration AD setting, as summarized in \cref{tab:ablation_ad}. Importantly, we keep all hyperparameters strictly unified across all scenes, avoiding any per-scene hyperparameters tuning. Since Mini-Splatting2~\cite{fang2024mini2} natively uses an AD-like strategy, we keep its default settings. We configure Taming-3DGS~\cite{mallick2024taming} to reach about 1M parameters, and relax the densification threshold of FastGS~\cite{ren2025fastgs} to achieve a similar 1M scale. Under this rapid densification, Taming-3DGS fails to converge because its density metrics cannot handle massive splitting in a short time. FastGS shows a clear speedup but suffers a $\sim$0.5 dB drop in PSNR, as it generates many invalid points without an efficient way to remove them. Furthermore, Mini-Splatting2 relies heavily on a long pruning phase; without it, its training time would significantly increase. In contrast, LiteGS easily supports this aggressive parameter growth, directly translating the reduced iterations into actual time savings while maintaining high rendering quality.

\begin{table}[t]
\centering
\caption{Ablation study on Aggressive Densification (AD). We compare standard configurations against AD adaptations (10K iterations, $\sim$1M parameters) under strictly unified hyperparameters across all scenes. Results show average metrics on the Mip-NeRF 360 dataset.}
\label{tab:ablation_ad}
\scriptsize
\begin{tabular}{l l c c c c c c}
\toprule
\textbf{Method} & \textbf{Config} & \textbf{Iters} & \textbf{Num (M)} & \textbf{Time (s)} & \textbf{SSIM} $\uparrow$ & \textbf{PSNR} $\uparrow$ & \textbf{LPIPS} $\downarrow$ \\
\midrule
Mini-Splatting2 \cite{fang2024mini2} & AD Default & 18K & 0.62 & 214 & 0.821 & 27.33 & 0.215 \\
\midrule
\multirow{2}{*}{Taming-3DGS \cite{mallick2024taming}} & Default & 30K & 0.68 & 380 & 0.801 & 27.32 & 0.257 \\
                             & + AD    & 10K & 1.0  & - & - & - & - \\
\midrule
\multirow{2}{*}{FastGS \cite{ren2025fastgs}}      & Default & 30K & 0.40 & 215 & 0.797 & 27.56 & 0.261 \\
                             & + AD    & 10K & 1.3 & 106 & 0.805 & 27.00 & 0.237 \\ 
\midrule
\multirow{2}{*}{LiteGS}      & Default & 30K & 0.42 & 127 & 0.804 & 27.62 & 0.250 \\
                             & + AD    & 10K & 1.0 & 64 & 0.813 & 27.60 & 0.229 \\
\bottomrule
\end{tabular}
\end{table}

\noindent\textbf{Impact of Cluster-Cull-Compact}. To isolate its effect, we disable our spatial clustering pipeline in the LiteGS-quality setting. The results in \cref{tab:ablationsccc} show that it significantly increases training time (e.g. 36\% slower, from 194s to 264s on Mip-NeRF 360) with only a minor impact on quality metrics. This reveals that our Cluster-Cull-Compact module effectively accelerates training.

\begin{table*}[t]
\centering
\scriptsize
\setlength{\tabcolsep}{1pt}
\renewcommand{\arraystretch}{1.1}

\caption{Ablation study on the effect of our Cluster-Cull-Compact pipeline. Disabling cluster-level culling (w/o culling) significantly increases training time with negligible impact on final image quality across all datasets.}
\label{tab:ablationsccc}

\begin{tabular}{l|cc|cc|cc}
\toprule
\multirow{2}{*}{\textbf{}} & \multicolumn{2}{c|}{\textbf{Mip-NeRF 360}} & \multicolumn{2}{c|}{\textbf{Tanks\&Temples}} & \multicolumn{2}{c}{\textbf{Deep Blending}} \\
\cmidrule(lr){2-3} \cmidrule(lr){4-5} \cmidrule(lr){6-7}
& SSIM/PSNR/LPIPS & Time(s)
& SSIM/PSNR/LPIPS & Time(s)
& SSIM/PSNR/LPIPS & Time(s) \\
\midrule
w/ culling
& 0.827/28.14/0.207 & 194
& 0.861/24.61/0.161 & 150
& 0.911/30.36/0.241 & 128 \\
\midrule
w/o culling
& 0.826/28.15/0.207 & 264 
& 0.862/24.63/0.160 & 182
& 0.913/30.36/0.241 & 206 \\
\bottomrule

\end{tabular}
\end{table*}

\noindent\textbf{Impact of Warp-based Rasterizer}. An end-to-end training time comparison is insufficient to fairly evaluate rasterizer performance, as our densification strategy may generate more visible primitives. Therefore, we list the cost breakdown of the operators in \cref{tab:rasterizer_comparison}. Our warp-based rasterizer demonstrates a prominent speedup across all primitive scales, outperforming the original 3DGS atomic rasterizer by 7.0x-13.0x, popular per-splat rasterizers by ~4.0x, and even recent SOTA methods~\cite{TCGS} using Tensor Cores by ~2x. 

\begin{table}[t]
    \begin{minipage}{0.51\textwidth}
        \centering
        \scriptsize
        \setlength{\tabcolsep}{1pt}
            \caption{Performance comparison of different rasterization methods on the `garden' scene with varying numbers of primitives (500k, 2000k, 5728k). Our Warp-based Rasterizer shows a significant speedup in both forward and backward passes. Atomic Raster is the original implementation in 3DGS \cite{kerbl20233d}. Per-splat Raster is the implementation from Taming-3DGS \cite{mallick2024taming} which may be the most popular gaussian raster operator. Tensor-core GS \cite{TCGS} is the recent SOTA work for 3DGS inference and Warp-based Raster is our work.}
            \label{tab:rasterizer_comparison}

            \begin{tabular}{l|ccc|ccc}
                \toprule
                 & \multicolumn{3}{c|}{\textbf{Forward(ms)}} & \multicolumn{3}{c}{\textbf{Backward(ms)}}\\
                \cmidrule(lr){2-4} \cmidrule(lr){5-7}
                & 500k & 2000k & 5728k & 500k & 2000k & 5728k \\
                \midrule
                Atomic & 5.45 & 10.56 & 17.82 & 25.26  & 37.23 & 51.84\\
                Per-splat & 5.06 & 7.49 &  12.58 & 7.53  & 13.46 & 27.31\\
                Tensor-core & 1.79 & 3.59 & 6.81 & -  & - & -\\
                Warp-based & \textbf{0.36} & \textbf{1.45} & \textbf{3.01} & \textbf{1.87}  & \textbf{3.64} & \textbf{6.93}\\
                
                \bottomrule
            \end{tabular}
            
    \end{minipage}\hfill
    \begin{minipage}{0.47\textwidth}
        \centering
        \scriptsize
        \caption{Ablation study on our densification strategy. We compare our full model against variants without opacity decay (-Decay), without the gradient variance metric (-Var), and without both (-Both).}
        \label{tab:densify_albation}
        \begin{tabular}{l|cccc}
            \toprule
             & LiteGS & -Decay & -Var & -Both \\
            \midrule
            \multicolumn{5}{c}{Mip-NeRF 360} \\
            \midrule
            PSNR & \textbf{28.14} & 27.68 & 27.40 & 27.13 \\
            SSIM & \textbf{0.827} & 0.813 & 0.813 & 0.777 \\
            LPIPS& \textbf{0.207} & 0.222 & 0.235 & 0.259 \\
            \midrule
            \multicolumn{5}{c}{Tanks\&Temples} \\
            \midrule
            PSNR & \textbf{24.61} & 24.11 & 24.27 & 23.97 \\
            SSIM & \textbf{0.861} & 0.852 & 0.853 & 0.837 \\
            LPIPS& \textbf{0.161} & 0.166 & 0.163 & 0.190 \\
            \midrule
            \multicolumn{5}{c}{Deep Blending} \\
            \midrule
            PSNR& \textbf{30.36} & 29.93 & 30.08 & 29.76 \\
            SSIM& \textbf{0.911} & 0.908 & 0.905 & 0.900 \\
            LPIPS& 0.241 & \textbf{0.239} & 0.245 & 0.259 \\
            \bottomrule
        \end{tabular}

    \end{minipage}
    
\end{table}

\noindent\textbf{Impact of Densification Strategy}. \cref{tab:densify_albation} depicts the performance of each component in our densification strategy. In the setting without opacity decay, we use the hard opacity reset scheme and the same hyperparameters in the original 3DGS. In the setting without opacity gradient variance metric, per-pixel error~\cite{Revising3DGS} is taken as the densification indicator.
Using the original 3DGS densification hyperparameters and opacity reset policy could drop PSNR by 0.4-0.5 dB. This suggests that a more frequent reduction in overall scene opacity improves reconstruction quality, and our opacity decay mechanism avoids the instability of consecutive hard resets by allowing faster recovery. In the control experiment using per-pixel error, the PSNR drops by approx. 0.3-0.7 dB, indicating that the opacity gradient variance is a more robust metric than the error-based indicator.

\section{Conclusion}
\label{sec:conclusion}

In this paper, we presented LiteGS, a high-performance framework that systematically accelerates 3D Gaussian Splatting training through fundamental system-algorithm codesign. By breaking the conventional isolation between hardware execution and algorithmic design, we introduced a warp-based rasterization paradigm, a Cluster-Cull-Compact data pipeline, and a robust variance-guided densification metric. This powerful synergy allows LiteGS to consistently dominate both training efficiency and rendering quality on all parameter scales. 

Extensive evaluations validate our milestone performance across distinct training regimes. In the high-quality track, LiteGS achieves at least a 10$\times$ speedup over the current quality SOTA while delivering matching or superior fidelity with significantly fewer parameters. Furthermore, by unlocking a highly effective ``space-for-time'' aggressive densification strategy (LiteGS*), we successfully compressed the training time to a sub-minute milestone without sacrificing competitive rendering quality. We believe LiteGS sets a new performance benchmark and provides a highly robust, flexible, and open-source foundation for future research in real-time novel view synthesis.

%
%
\bibliographystyle{splncs04}
\bibliography{main}

\clearpage
\appendix

\section{Detail of GS Forward Differencing} \label{appendix:scaneline}

For Gaussian primitives, consider the computation of 2D Gaussian rasterization in \cref{equ:gaussian}, where $\Sigma^{\text{screen}} = \begin{pmatrix} a & b \\ b & c \end{pmatrix}$. \cref{equ:gaussian} can be expanded into \cref{equ:gaussian_abc}.

\begin{equation}
\label{equ:gaussian_abc}
G(\Delta x, \Delta y, \Sigma) = e^{-0.5(a\Delta x^2 + 2b\Delta x \Delta y +c\Delta y^2)}
\end{equation}

Considering the $i$-th point $(x,y+i)$ on the scanline, $G(x^{screen}-x,y^{screen}-(y+i),\Sigma^{screen})$ can be expressed as \cref{equ:gaussian_scanline_abc}.

\begin{equation}
\label{equ:gaussian_scanline_abc}
G(\Delta x, \Delta y - i, \Sigma) = e^{-0.5(a\Delta x^2 + 2b\Delta x \Delta y +c\Delta y^2) + (b\Delta x + c\Delta y)i - 0.5c i^2}
\end{equation}

By defining $\text{Basic}=-0.5(a\Delta x^2 + 2b\Delta x \Delta y +c\Delta y^2)$, $\text{Linear}=(b\Delta x + c\Delta y)$, and $\text{Quad}=-0.5c$, we obtain the scanline computation formula for Gaussian primitive software rasterization in \cref{equ:gaussian_scanline}. 

\begin{equation}
\label{equ:gaussian_scanline}
G(\Delta x, \Delta y -i, \Sigma) = e^{\text{Basic} + \text{Linear} \cdot i + \text{Quad} \cdot i^2}
\end{equation}

The warp-based rasterization approach for gradient summation also benefits from the scanline. For parameter $c$, let $l=\sum_{i=0}^{L} \frac{\partial Loss}{\partial I(x,y+i)} \frac{\partial I(x,y+i)}{\partial c}$. The per-pixel gradient computation is shown in \cref{equ:c_grad}. The sum of gradients along the scanline is expressed in \cref{equ:c_grad_scan}.

\begin{equation}
\scriptsize
\begin{aligned}
\label{equ:c_grad}
l&=\sum_{i=0}^{L} \frac{\partial Loss}{\partial I(x,y+i)} \frac{\partial I(x,y+i)}{\partial G(\Delta x,\Delta y -i, \Sigma)} \frac{\partial G(\Delta x,\Delta y -i, \Sigma)}{\partial c}\\
&=\sum_{i=0}^{L} \frac{\partial Loss}{\partial I(x,y+i)} \frac{\partial I(x,y+i)}{\partial G(\Delta x,\Delta y -i, \Sigma)} G(\Delta x,\Delta y -i, \Sigma) \cdot (-0.5\Delta y^2 + \Delta y i - 0.5 i^2)
\end{aligned}
\end{equation}

\begin{equation}
\scriptsize
\begin{aligned}
\label{equ:c_grad_scan}
l &= \frac{\partial \text{Basic}}{\partial c} \sum_{i=0}^{L} \frac{\partial Loss}{\partial I(x,y+i)} \frac{\partial I(x,y+i)}{\partial G(\Delta x,\Delta y -i, \Sigma)} \frac{\partial G(\Delta x,\Delta y -i, \Sigma)}{\partial \text{Basic}}\\
&+\frac{\partial \text{Linear}}{\partial c} \sum_{i=0}^{L} \frac{\partial Loss}{\partial I(x,y+i)} \frac{\partial I(x,y+i)}{\partial G(\Delta x,\Delta y -i, \Sigma)} \frac{\partial G(\Delta x,\Delta y -i, \Sigma)}{\partial \text{Linear}}\\
&+\frac{\partial \text{Quad}}{\partial c} \sum_{i=0}^{L} \frac{\partial Loss}{\partial I(x,y+i)} \frac{\partial I(x,y+i)}{\partial G(\Delta x,\Delta y -i, \Sigma)} \frac{\partial G(\Delta x,\Delta y -i, \Sigma)}{\partial \text{Quad}}
\end{aligned}
\end{equation}

\section{Detail of Cluster-Cull-Compact} \label{appendix:cluster-cull-compact}

\begin{figure}
  \centering
  \includegraphics[width=0.6\columnwidth]{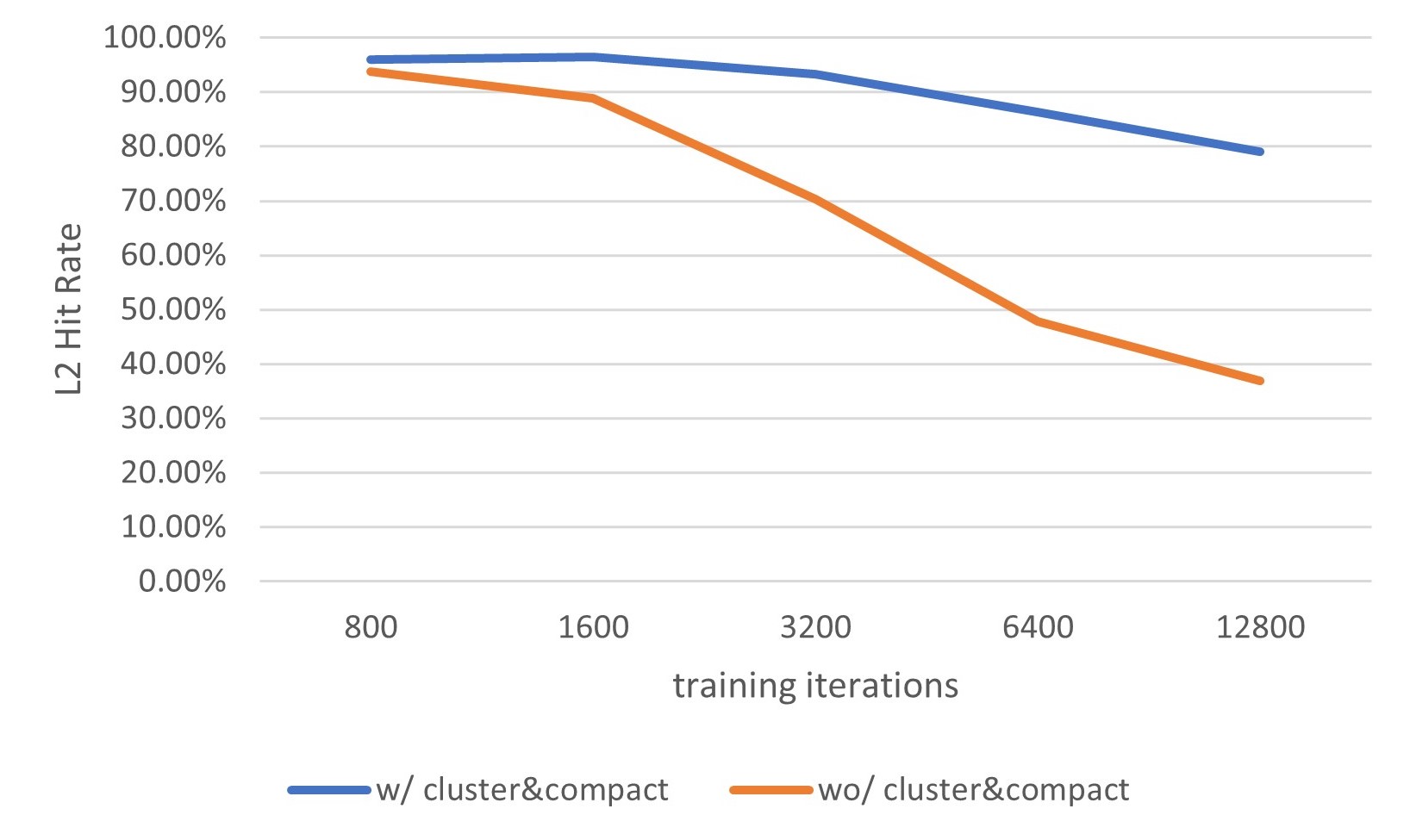}
  \caption{The evolution of L2 cache hit rate throughout training. The proposed cluster-and-compact method (blue) maintains significantly higher cache efficiency compared to the baseline without this optimization (orange), as the model grows and primitives are appended.}
  \label{image:hitrate}
\end{figure}

The 3DGS training phase mainly faces two major issues: (1) \textbf{cache misses} caused by the unordered storage of scene primitives, and (2) \textbf{hardware inefficiency} issues introduced by view frustum culling. \cref{image:hitrate} illustrates the change in the GPU L2 cache hit rate throughout the training process as the model gradually expands. The figure clearly shows that as training progresses, new parameters are appended to the end of the memory space, degrading the coherence between the 3D spatial structure and the memory layout. Consequently, the L2 cache hit rate progressively decreases during training.

\begin{figure}
  \centering
  \includegraphics[width=0.7\columnwidth]{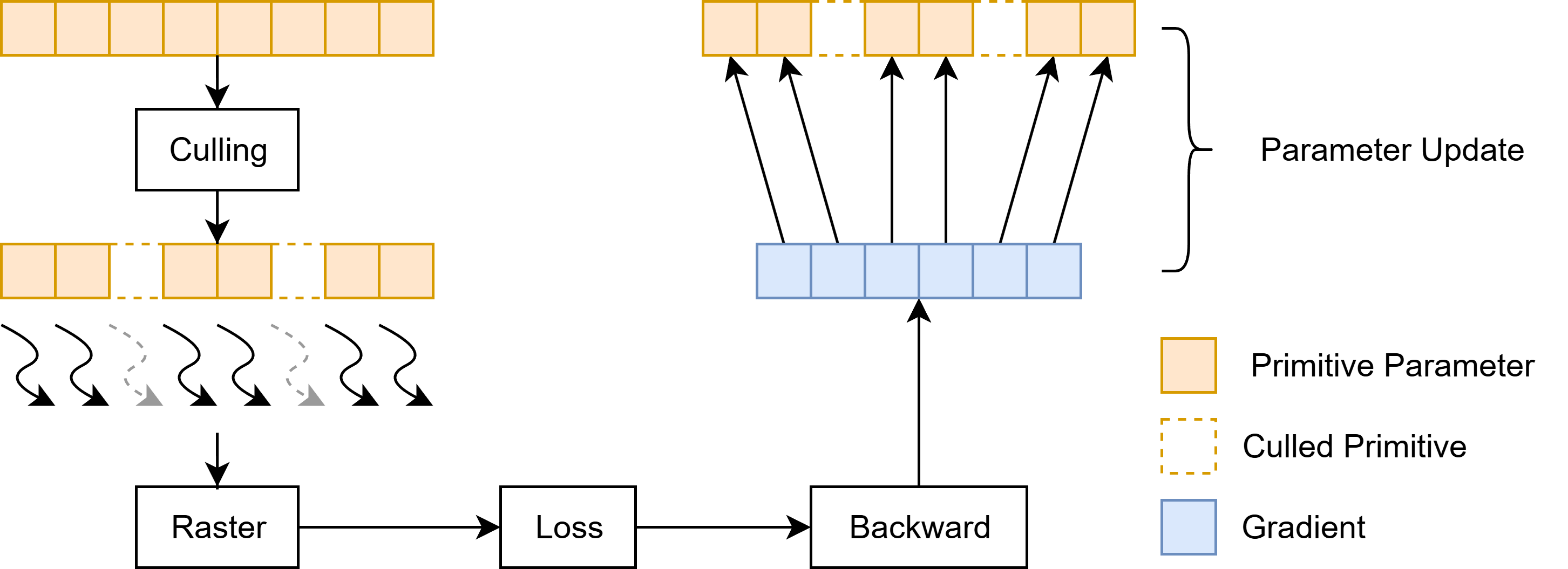}
  \caption{Illustration of GPU inefficiency caused by culling. Primitives are stored unordered in memory and view frustum culling removes primitives arbitrarily. This leads to warp divergence, as well as inefficient memory operations during the parameter update with sparse gradients.}
  \label{image:divergence}
\end{figure}

\cref{image:divergence} demonstrates the computational and I/O efficiency problems caused by culling. As primitives lose the coherence between their 3D spatial location and their memory location, culling operations can occur randomly at any location within the memory space. Since GPU computation is scheduled in warps ($32$ threads), primitive-related computations are only skipped if all $32$ consecutively aligned primitives within a warp are culled. In the vast majority of cases, culled primitives do not actually reduce the number of instructions executed. A similar issue occurs during parameter updates when using sparse gradients. Global memory load/store operations on the GPU are performed at a minimum granularity of a 32-byte sector. When threads attempt to write to several noncontiguous addresses, this often results in the hardware storing multiple entire sectors. In this scenario, sparse gradients do not truly reduce the amount of data written to memory.

\section{Qualitative Results} 
\label{section:visual}

\begin{figure}[ht]
\centering

\setlength{\tabcolsep}{3pt}

\begin{tabular}{ccc}  
GT & 3DGS-MCMC & LiteGS \\
\includegraphics[width=0.3\linewidth]{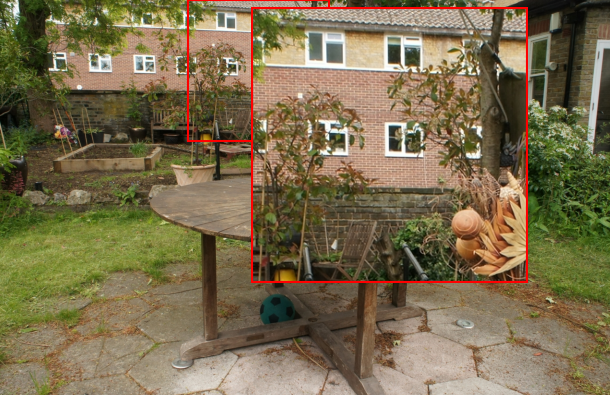}
& \includegraphics[width=0.3\linewidth]{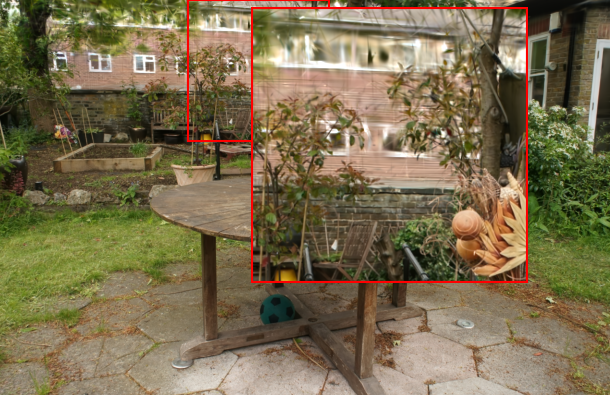}
& \includegraphics[width=0.3\linewidth]{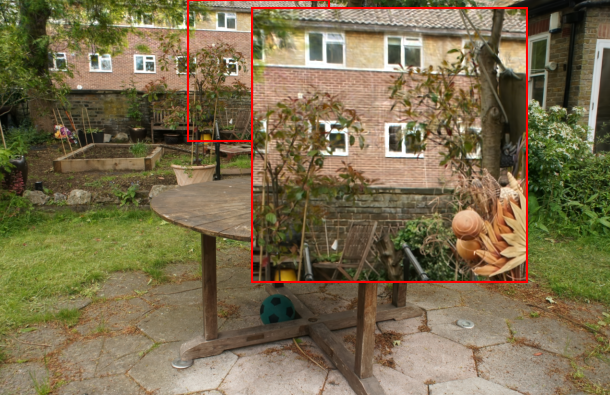} \\

\includegraphics[width=0.3\linewidth]{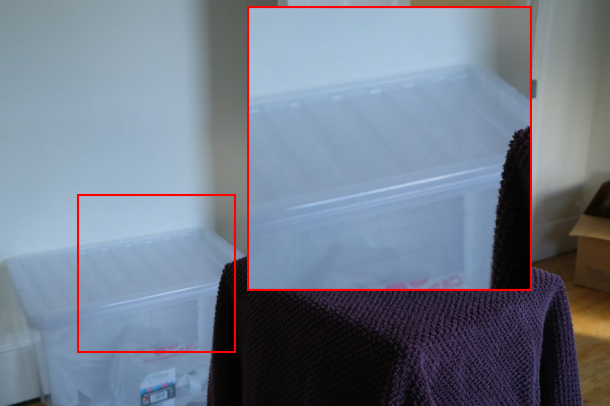}
& \includegraphics[width=0.3\linewidth]{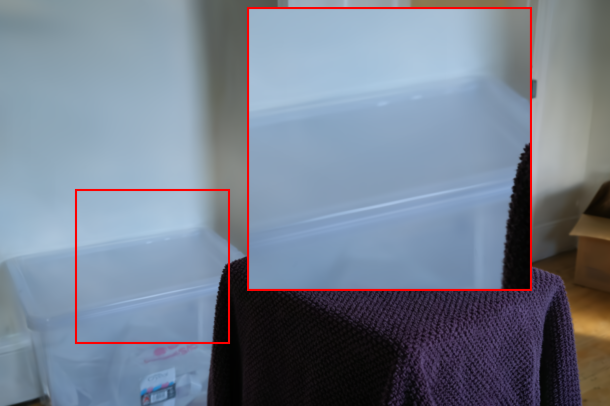}
& \includegraphics[width=0.3\linewidth]{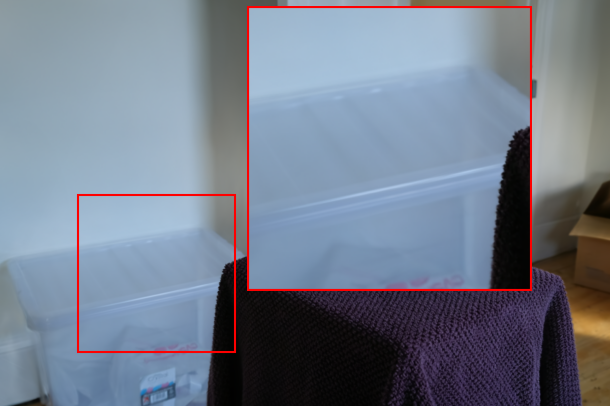} \\

\includegraphics[width=0.3\linewidth]{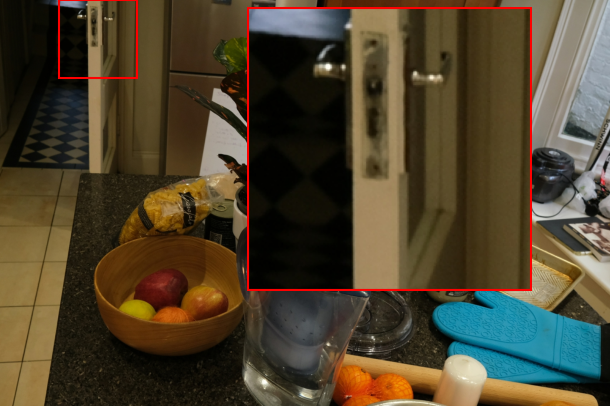}
& \includegraphics[width=0.3\linewidth]{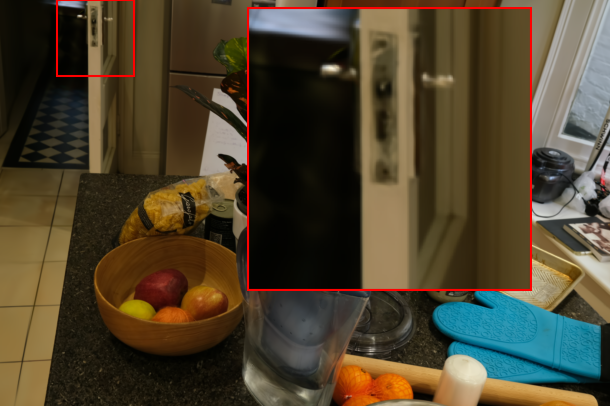}
& \includegraphics[width=0.3\linewidth]{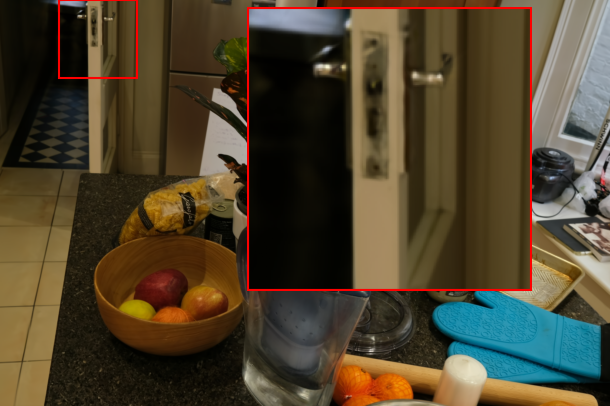} \\
\end{tabular}
\caption{Qualitative comparisons between LiteGS and 3DGS-MCMC. GT denotes ground truth.}
\label{fig:imagegrid}
\end{figure}



\end{document}